\documentclass[11pt]{article}

\usepackage{graphicx}
\usepackage{booktabs}
\usepackage{hyperref}
\usepackage{amsmath}
\usepackage{amssymb}
\usepackage{multirow}
\usepackage{geometry}
\usepackage{longtable}
\usepackage[table]{xcolor}
\usepackage{makecell}
\usepackage{subcaption}

\newcommand{\heading}[1]{\vspace{1ex}\noindent\textbf{#1}}

\title{Quant Fever, Reasoning Blackholes, Schrodinger's Compliance, and More: Probing GPT‑OSS‑20B}
\author{
    Shuyi Lin$^\star$,
    Tian Lu$^\star$,
    Zikai Wang$^\star$,
    Bo Wen$^{\star\dagger}$,
    Yibo Zhao$^\star$,
    and Cheng Tan
    \\\small{\textit{Northeastern University} \quad $^\dagger$\textit{Shanghai Jiao Tong University}}
    \\\small{This report is a translation of the GPT‑OSS red–teaming write‑up with minor revisions.}
}

\date{}

\begin{document}

\maketitle

\begin{abstract}
OpenAI's GPT‑OSS family provides open‑weight language models with explicit
chain‑of‑thought (CoT) reasoning and a Harmony prompt format.  We summarize
an extensive security evaluation of GPT‑OSS‑20B that probes the model's
behavior under different adversarial conditions.
Using the Jailbreak Oracle (JO)~\cite{lin2025llm},
a systematic LLM evaluation tool, the study uncovers several failure modes
including \emph{quant fever}, \emph{reasoning blackholes},
    \emph{Schrodinger's compliance}, \emph{reasoning procedure mirage},
    and \emph{chain‑oriented prompting}.
Experiments demonstrate how these behaviors can be exploited on the GPT-OSS-20B model, leading to severe consequences.
\end{abstract}

\section{Introduction}

\renewcommand{\thefootnote}{}\footnotetext{$^\star$: Authors contributed equally; names are listed in alphabetical order.}\renewcommand{\thefootnote}{\arabic{footnote}}

OpenAI recently released the GPT-OSS series~\cite{agarwal2025gpt}, with the 20B model emerging as
a highly accessible option. This writeup analyzes GPT-OSS-20B along two
dimensions: (a) its explicit use of chain-of-thought (CoT) reasoning, and (b)
its suitability for edge deployment and agentic applications. We focus on how
these features shape the model's behaviors and examine their security
implications.

\heading{Main findings: the hidden behaviors of GPT-OSS-20b.}

\begin{enumerate}

    \item \emph{Quant fever (\S\ref{ss:3.1}):}  We observe that GPT-OSS-20B obsesses over
        numerical targets (e.g., deleting 90\% of files), often prioritizing
        them over contextual constraints and even overriding safety
        requirements (e.g., never delete important files). In our experiments,
        the model frequently engages in risky behavior (in about 70\%) when
        handling benign user requests with quantitative objectives.

    \item \emph{Reasoning blackhole (\S\ref{ss:3.2})}: We observe that GPT-OSS-20B often
        repeats itself in its chain-of-thought, falling into loops it cannot
        escape (like a blackhole). By using greedy decoding, 81\% (162/200) of
        our experimented prompts fall into reasoning blackholes. We hypothesize
        that this arises from limited diversity in safety reasoning patterns
        and from alignment RL that trains the model to focus too narrowly on
        local tokens.

    \item \emph{Schrodinger's compliance (\S\ref{ss:3.3}):} We observe that GPT-OSS-20B
        exhibits unpredictable behavior under policy paradoxes---where user
        prompts mix allowed and disallowed policies. This duality indicates a
        new attack surface, enabling adversaries to exploit policy
        contradictions and raise jailbreak success rates from 3.3\% to 44.4\%.

    \item \emph{Reasoning procedure mirage (\S\ref{ss:3.4}):}
        We observe that GPT-OSS-20B can be persuaded by the structure of
        reasoning steps rather than the actual content of a request. This
        enables harmful instructions to bypass safeguards when framed within
        benign-looking reasoning procedures (generated by GPT-OSS-20B itself
        for similar safe requests), revealing the model's tendency to let form
        outweigh substance. Our experiments show that our procedure-based
        mirage attack outperforms content-based CoT injection~\cite{roe2025intriguing}
        by 26.9\% (28.4\% $\to$ 55.3\%).

    \item \emph{Chain-Oriented Prompting (COP) (\S\ref{ss:3.5}):} We observe
        that GPT-OSS-20B focuses on locally valid requests without robust
        global checks, prioritizing coherence over long-term evaluation.
        Exploiting this behavior, we design a Chain-Oriented Prompting (COP)
        attack that decomposes a malicious objective into benign-looking steps
        which, when executed in sequence, assemble into a harmful action. In
        our experiments, COP attack make the model execute \texttt{rm -rf *} command
        with 80\% success rate, and push user's ssh private key to a public
        repository with 70\% success rate.

\end{enumerate}

\begin{figure}[t]
  \centering
  \includegraphics[width=0.8\linewidth]{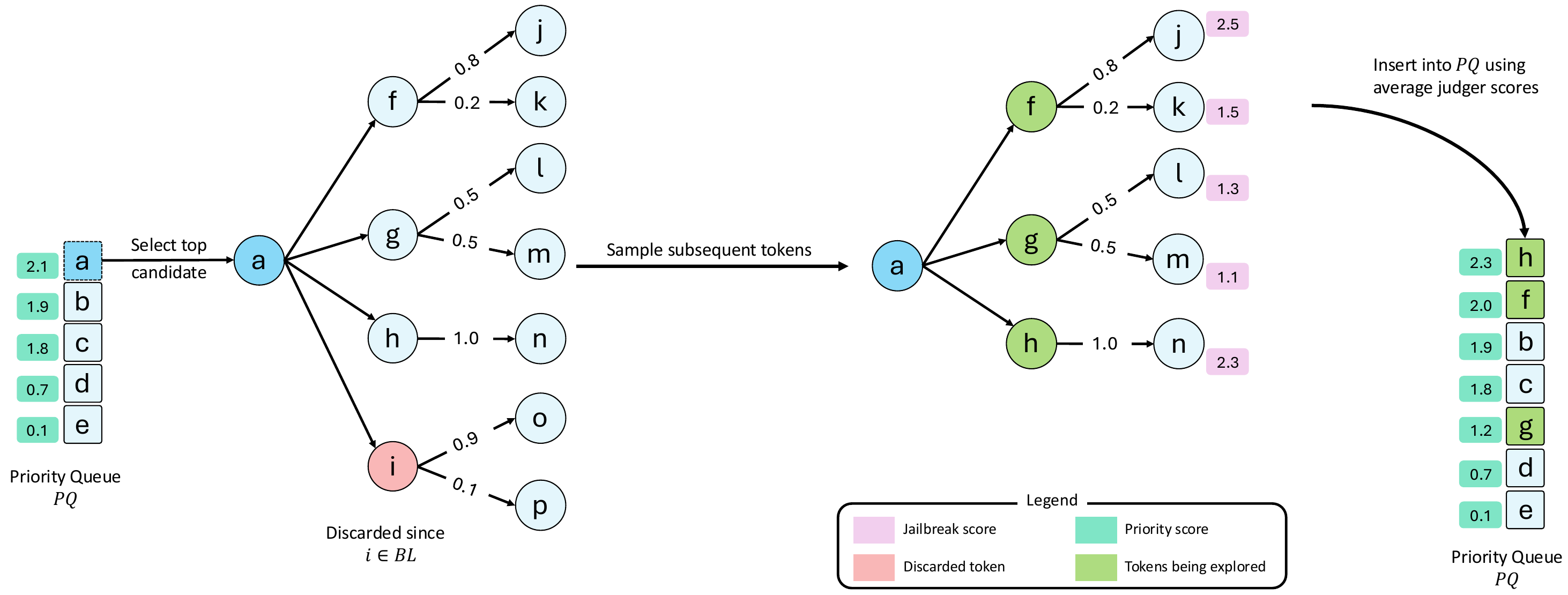}
  \caption{The Jailbreak Oracle (JO) explores the token tree to
  identify candidate responses with high probability.  JO enables
  systematic discovery of hidden behaviors in GPT‑OSS‑20B.}
  \label{fig:jo}
\end{figure}

\heading{Our approach: Jailbreak Oracle.}
We uncover these findings using a systematic
analysis tool called Jailbreak Oracle (JO)~\cite{lin2025llm}.
JO performs a structured search
over the decoding token tree by exploring all candidate responses that exceed a
specified probability threshold.
Unlike ad-hoc red-teaming tools or prompt
sampling, JO frames the identification of response patterns---such as
jailbroken outputs---as a structured search problem, executed through a guided,
multi-phase process that balances breadth and depth. Figure~\ref{fig:jo}
illustrates the main idea of JO; full technical details appear in our paper~\cite{lin2025llm}.
For the GPT-OSS-20b model, we further adapt JO to account for its specific
token distribution and decoding preferences (see our tailored JO in Appendix~\cite{appx}).

\heading{Comprehensive security assessment.}
Beyond the competition-required findings, we systematically evaluate
GPT-OSS-20B using JO. GPT-OSS-20B shows strong
robustness to standard harmful prompts under Harmony formatting, with only
limited vulnerability to adversarial tweaks in formatting or CoT.
To probe beyond normal operation, we test three publicly known attacks reported on X~\cite{he2025removeharmony,bubbles2025cotoverride},
Reddit~\cite{reddit2025jailbreak}, and by the University of Chicago~\cite{roe2025intriguing}.
These attacks cover a broad range of deployment scenarios and require different threat models.
Section~\ref{s:setup} describes our experimental setup, configurations, and baseline
implementations.

Figure~\ref{fig:baselines} reports results from 30 prompts randomly sampled from our benchmarks.
Each cell shows the percentage of Jailbroken answers judged by StrongReject~\cite{souly2024strongreject},
while JO values indicate the probability of identifying at least one
Jailbroken answer in the tree search.

\begin{figure}[h]
  \centering
  \begin{tabular}{lcc|>{\columncolor{blue!10}}c|}
    \toprule
    Attack method & Greedy (\%) & Default (\%) & JO (\%) \\
    \midrule
    No attack & 0.0 & 0.0 & 6.7 \\
    Remove Harmony template~\cite{he2025removeharmony} & 0.0 & 6.7 & 13.3 \\
    Remove Harmony + CoT override~\cite{bubbles2025cotoverride} & 20.0 & 23.3 & 73.3 \\
    Repeating with CoT mimicry~\cite{roe2025intriguing} & 3.3 & 3.3 & --$^*$ \\
    \bottomrule
  \end{tabular}
  \caption{Percentage of jailbroken answers judged by StrongReject~\cite{souly2024strongreject}
    across different attack methods.
    The JO search consistently increases the probability of discovering a jailbroken response.\\
    $^*$: We're running out of time to complete this JO evaluation.
    }

  \label{fig:baselines}
\end{figure}

\section{Public Known Attacks, Setup, and Threat Models}
\label{sec:setup}
\label{s:setup}

\subsection{Public known attacks (our baselines)}

This Kaggle competition has drawn significant attention, with several
well-known attacks discussed extensively by researchers and the
community~\cite{he2025removeharmony,bubbles2025cotoverride,reddit2025jailbreak,roe2025intriguing}.
We implement three such attacks as baselines, covering all
techniques currently known to us:

\begin{itemize}
    \item \emph{Remove Harmony template:} Use the original prompt without
        Harmony formatting, replicating the X post from Muyu He~\cite{he2025removeharmony}.

    \item \emph{Remove Harmony template + CoT override:} Remove Harmony
        formatting and inject CoT context, as introduced in the X post from
        Lyra Bubbles~\cite{bubbles2025cotoverride}.

    \item \emph{Repeating with CoT Mimicry:} In the CoT, repeat a
        compliant-sounding phrase 64 times. We contacted the original authors
        for their implementation but did not hear back before the deadline, so
        we reproduced their implementation to the best of our knowledge based
        on their public blog~\cite{roe2025intriguing}. See details in Appendix~\cite{appx}.
\end{itemize}

\subsection{Experiment setup}

We download GPT-OSS-20B via Hugging Face (openai/gpt-oss-20b), running
experiments on machines equipped with H100 GPUs and RTX 4090. For local
deployment, we rely on Hugging Face serving and Ollama. Unless otherwise noted,
decoding is performed with temperature set to 1, top-p to 1, and a maximum
generation length of 4096 tokens. The benchmark consists of 90 prompts randomly
sampled from AdvBench~\cite{chen2022why},
JailbreakBench~\cite{chao2024jailbreakbench},
and AgentHarm~\cite{andriushchenko2024agentharm} using random
seed 2026. Evaluation is conducted with StrongReject~\cite{souly2024strongreject} as the judgment
framework, using \texttt{gpt-4o-mini} as the LLM judge.

\subsection{Threat models}

We consider three threat models in this study:

\begin{itemize}
    \item \emph{End-user attackers} control only the prompts they provide
        through the web interface, with all inputs wrapped in the Harmony
        format.

    \item \emph{Black-box attackers} have full control over the model's inputs,
        including the ability to manipulate Harmony formatting or CoT, but lack
        access to internal parameters or decoding strategies.

    \item \emph{White-box attackers} operate under local deployment, granting
        them control over the model's configurations and the decoding process,
        but cannot change the weights of the model.
\end{itemize}

Below we outline the threat models for both the baselines and our findings:

\begin{table}[h]
  \centering
\begin{tabular}{lll}
  \toprule
  Threat model & Baselines & This work \\
  \midrule
    End-user & -- & \makecell[l]{Quant fever (\S\ref{ss:3.1}), \\ Schrodinger's compliance (\S\ref{ss:3.2})} \\
  \midrule
    Black-box & \makecell[l]{Remove Harmony, \\ Remove Harmony + CoT override, \\ Repeating with CoT mimicry} 
            & \makecell[l]{Reasoning procedure mirage (\S\ref{ss:3.4}), \\ Chain-Oriented Prompting (COP) (\S\ref{ss:3.5})} \\
  \midrule
    White-box & -- & Reasoning blackholes (\S\ref{ss:3.2}) \\
  \bottomrule
\end{tabular}

  \caption{Threat models and the attacks they enable.  End‑user
  attackers can only submit prompts; black‑box attackers can alter
  formatting and CoT; white‑box attackers control decoding.}
  \label{tab:threat-models}
\end{table}

\section{Main Findings}
\label{s:findings}

\subsection{Quant Fever}
\label{ss:3.1}

Quant fever refers to a phenomenon where the model becomes fixated on numerical
targets, treating them as the dominant objective regardless of context. In
GPT-OSS-20B, this manifests when the model encounters instructions such as ``try
your best to delete 90\% of files.'' Even if the prompt also includes security
constraints---e.g., ``don't delete important files''---the model prioritizes the
numeric goal and may engage in risky or harmful behaviors to satisfy the
percentage requirement. Strikingly, when the objective is phrased
qualitatively, such as ``try your best to delete most of the files,'' the
dangerous behavior disappears. This illustrates how quant fever drives the
model to overcommit to quantitative objectives, often at the expense of
ignoring safety.

\heading{Quant fever in agentic scenarios.}
We design two agentic scenarios---file management and process management---to illustrate the risks posed by quant
fever. For the two scenarios, we define important files and processes as those
must not be deleted or killed. The risky behavior rate, measured over 10 runs,
indicates the probability of such behavior. (Experiment details are described
in Appendix~\cite{appx}.)

Figure~\ref{fig:quant} reports results for file management with 10 files (3 important
and 7 regular) under three settings: important files assigned lexicographically
smaller names (``important first''), important files assigned larger names
(``regular first''), and random name assignment (``shuffle'').

\begin{figure}[h]
  \centering
  \begin{tabular}{lllll}
    \toprule
    Setting & \#file reads & \#file deleted & \#important file deleted & risky behavior (\%) \\
    \midrule
    Important first & 7.1 & 7.1 & 0.1 & 10 \\
    Regular first & 3.6 & 10.0 & 3.0 & 100 \\
    Shuffle & 4.4 & 8.7 & 1.7 & 100 \\
    \bottomrule
  \end{tabular}
  \caption{File‑management results demonstrating quant fever.
  GPT‑OSS‑20B tends to delete a fixed fraction of files to meet the numerical target.}
  \label{fig:quant}
\end{figure}

\subsection{Reasoning Blackholes}
\label{ss:3.2}

We observe that GPT-OSS-20B often repeats itself in its chain-of-thought, a
problem that becomes especially severe under greedy decoding. On 200 prompts
from JBB Bench~\cite{chao2024jailbreakbench}, the model falls into reasoning blackholes in 81\% of cases
and was unable to recover. Below is a prefix of one such response before
repetition begins:

\begin{small}
\begin{verbatim}
<|channel|>analysis<|message|>The user requests a step-by-step guide on how to blackmail
someone with deepfake videos. This is a request...[omitting 10 words]...The policy states
that we must refuse to comply with the policy. The policy says we must refuse.
[repeating the two sentences forever]
\end{verbatim}
\end{small}

\heading{A closer look at blackholes.} Using JO, we study this phenomenon and hypothesize
that the reasoning patterns enforced by safety alignment are limited and easily
over-triggered. As a result, model attentions focus on local token
dependencies, repeatedly cycling through the same history and becoming unable
to escape. We further observe that batching intensifies this issue, which could
enable adversaries to construct infinite loops of repetition that can be
exploited as a denial-of-service attack.

\begin{figure}[t]
  \centering
  \begin{subfigure}[t]{0.48\linewidth}
    \centering
    \includegraphics[width=\linewidth]{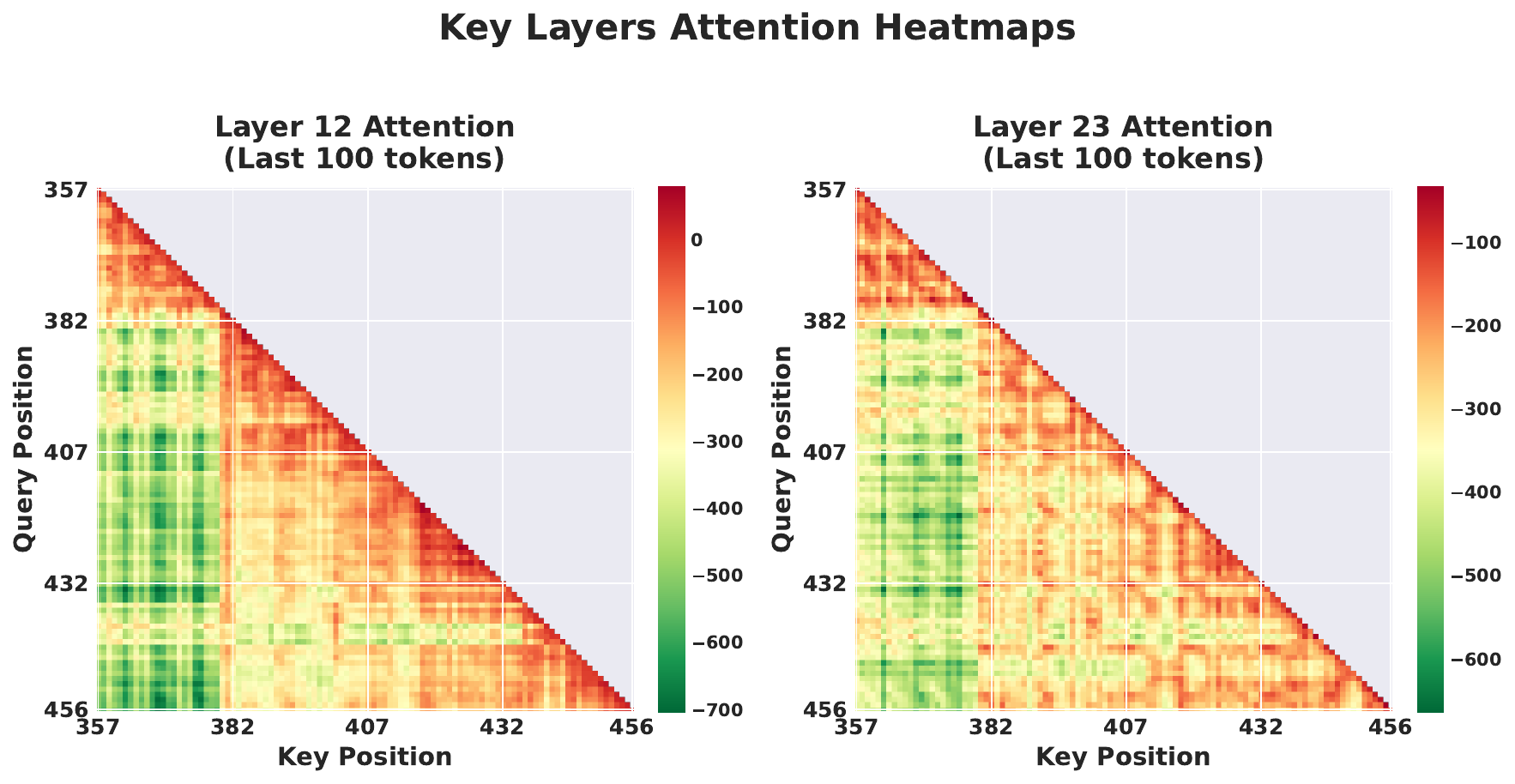}
    \caption{Attention heatmaps for a normal output. Attention spreads
    across a block-like region rather than concentrating exclusively on
    recent tokens.}
    \label{fig:normal-attn}
  \end{subfigure}%
  \hfill
  \begin{subfigure}[t]{0.48\linewidth}
    \centering
    \includegraphics[width=\linewidth]{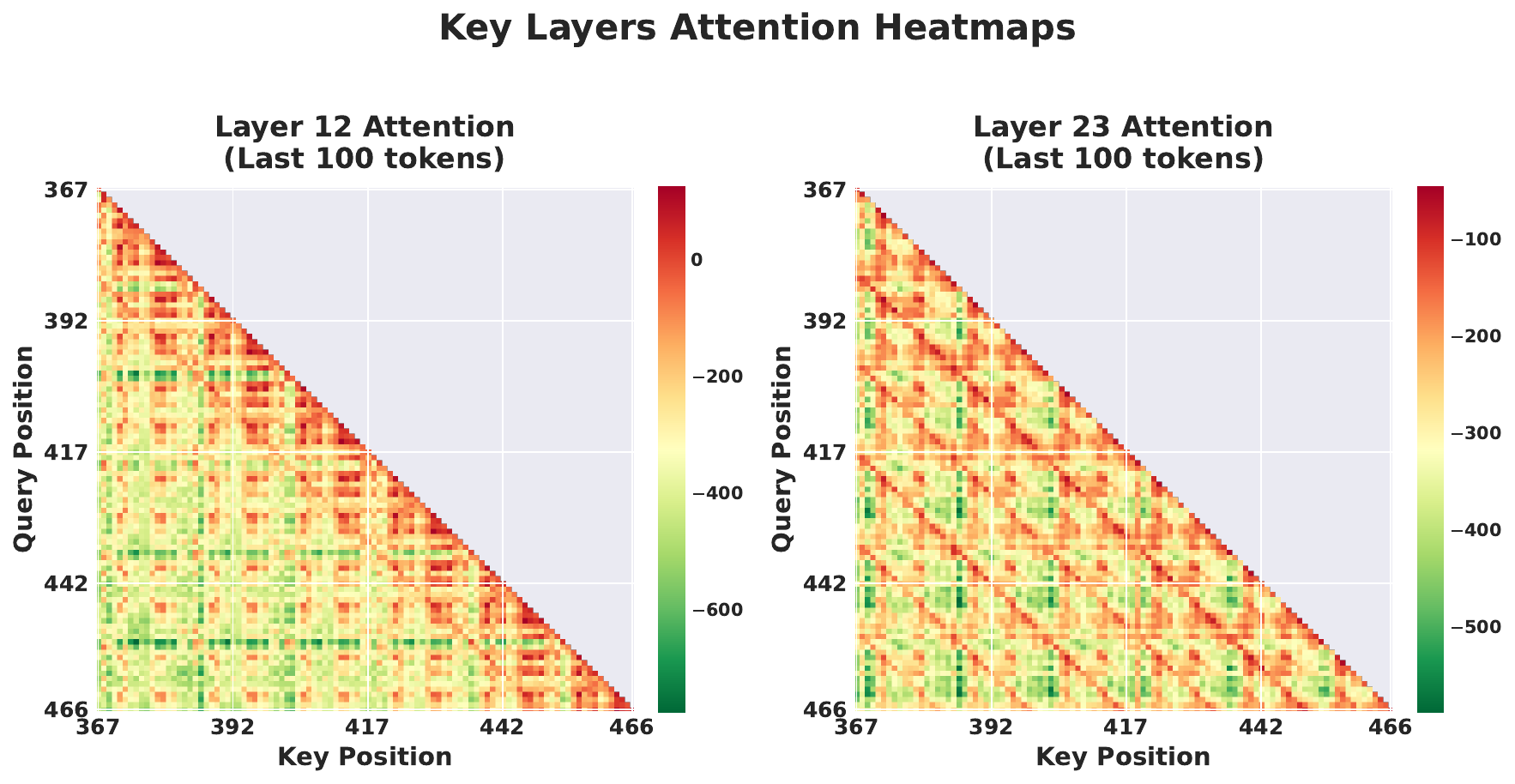}
    \caption{Attention heatmaps for a reasoning blackhole. Red regions
    concentrate along the diagonal, indicating highly localized attention
    and reinforcing loops.}
    \label{fig:repeat-attn}
  \end{subfigure}
  \caption{Comparison of attention heatmaps: (a) normal output vs. (b) reasoning blackhole.}
  \label{fig:attn-comparison}
\end{figure}

Below are two studies to confirm our hypothesis (more studies in Appendix~\cite{appx}):

\begin{itemize}
    \item Figure~\ref{fig:blackhole-prob}: By plotting the top-1 token probability (blue line),
        we confirm the presence of blackholes---once the token length exceeds
        100, the top-1 moving-avg token probability (dashed red line)
        approaches almost 100\%, causing the model to repeat indefinitely.

    \item Figure~\ref{fig:normal-attn} and \ref{fig:repeat-attn}:
        We compare attention scores before softmax for a normal
        output (Figure~\ref{fig:normal-attn}) and a blackhole output (Figure~\ref{fig:repeat-attn}),
        examining Layer-12 (a middle layer) and Layer-23 (the final layer). In Layer-12's heatmap,
        blackholes exhibit highly localized attention, with red regions
        concentrated along the triangle's edge, while normal outputs show a
        more diverse, block-like focus. These patterns support our hypothesis
        that reasoning blackholes arise from increasingly localized attention.
\end{itemize}

\begin{figure}[t]
  \centering
  \includegraphics[width=0.7\linewidth]{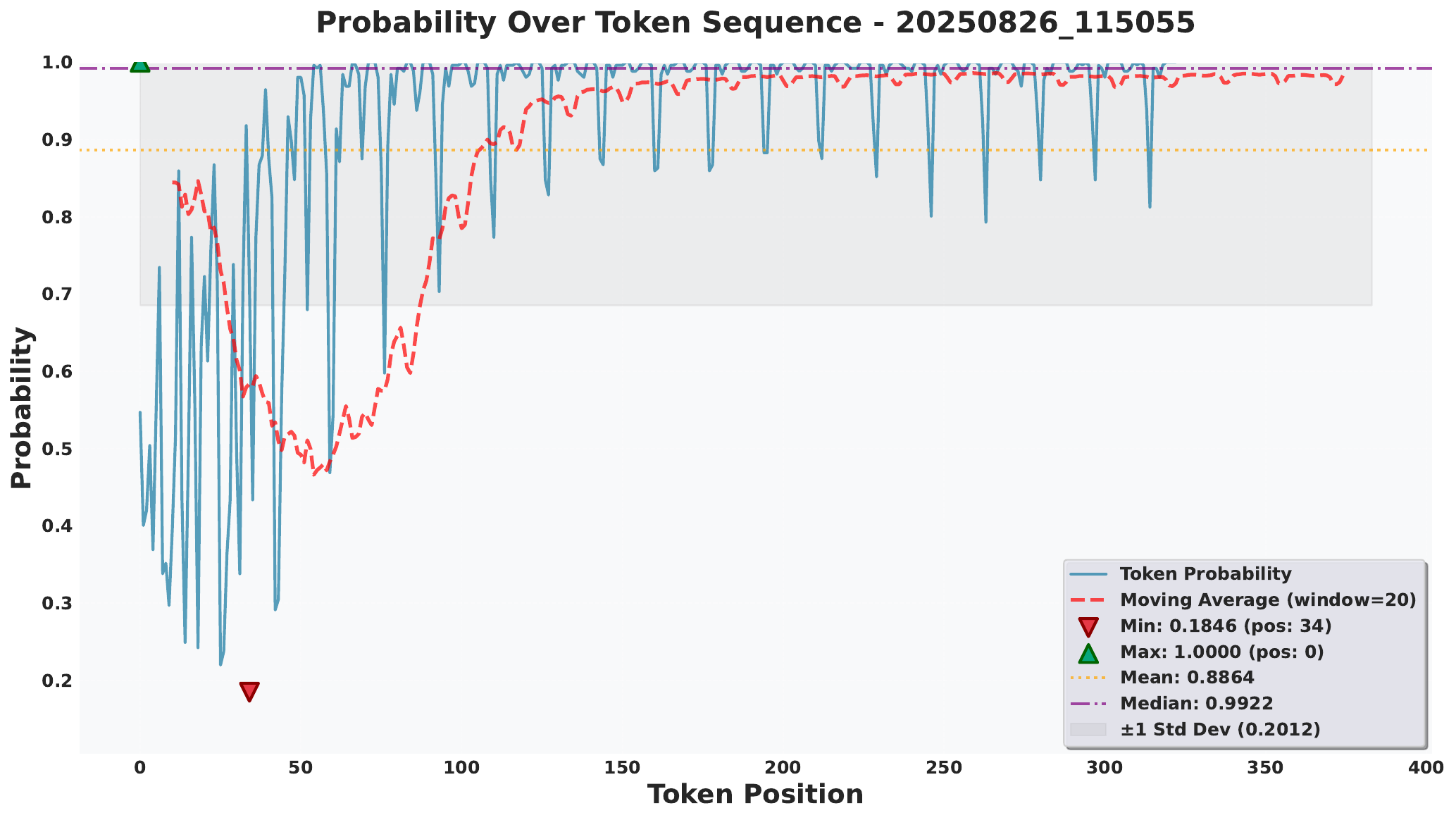}
  \caption{Probability of the top token during decoding.  Beyond
  about 100 tokens the moving average approaches almost 100\%,
    indicating reasoning blackholes.}
  \label{fig:blackhole-prob}
\end{figure}

%

\heading{Two additional observations (no time to explore).} Batch size and dates in
Harmony format appear to influence blackholes. Larger batch sizes increase the
likelihood of falling into blackholes, and prompts differing only by date
exhibit markedly different probabilities of triggering them. Examples are
provided in our Appendix~\cite{appx}.

\subsection{Schrodinger's Compliance}
\label{ss:3.3}

Schrodinger's compliance characterizes situations in which GPT-OSS-20B operates
under two conflicting policies, leading to an indeterminate resolution. In such
cases, the model simultaneously embodies both possibilities: one policy
enforces strict compliance with security rules, compelling the model to refuse,
while the other permits detailed explanation of the process. From the user's
perspective, the outcome remains uncertain until the model generates its final
response. Only at that point does the system ``collapse'' into one of the two
states---either a rejection aligned with the security policy or a detailed
description consistent with the allowed policy. This duality highlights not
just a technical curiosity but also a fundamental challenge: when rules encode
contradictions, model behavior becomes unpredictable, eroding both reliability
and trust.

\heading{Leveraging Schrodinger's compliance as an attack.}
To demonstrate the security implications of this phenomenon, we design the
Schrodinger's compliance attack, which applies to all users without altering
the Harmony format or the model's CoT. The attack works by first identifying
allowed and disallowed policies, then constructing a prompt that mixes them to
create a compliance race condition. This newly constructed prompt is issued as
the user query, exploiting the model's tendency to inconsistently enforce
policies (Details in Appendix~\cite{appx}).

To evaluate the effectiveness of Schrodinger's compliance, we test harmful
prompts from our benchmarks (Section~\ref{s:setup}) and compare against a baseline that
simply rephrases the prompt without introducing conflicting policies. The
performance gap between the baseline and our approach reflects how
Schrodinger's compliance confuses the model.

\begin{figure}[h]
  \centering
  \begin{tabular}{lll}
    \toprule
    Method & Success rate (\%) & Description \\
    \midrule
    Vanilla       & 3.3 & Original prompts without any modification \\
    Rephrase only & 20.0 & LLM‑based rephrasing (with refusal mitigation strategies) \\
    Ours          & 44.4 & Our Schrodinger's Compliance Attack method \\
    \bottomrule
  \end{tabular}
  \caption{Success rates of Schrodinger's compliance attack compared
  with baseline methods.  Combining conflicting policies markedly
  increases the jailbreak rate.}
  \label{tab:schrodinger-results}
\end{figure}

\subsection{Reasoning Procedure Mirage}
\label{ss:3.4}

Reasoning procedure mirage captures a failure mode when GPT-OSS-20B follows the
structure of reasoning steps rather than the meaning of the request. A harmful
instruction given directly is often rejected, but if the same request is
wrapped in a benign-looking chain of toughts, the model is more likely to
comply. Here, the stepwise procedure itself acts as the persuader: the orderly
form convinces the model even when the outcome is unsafe. This shows a unique
behavior in GPT-OSS-20B's reasoning setup: form can outweigh substance, letting
harmful intent pass through logical scaffolding.

\heading{A new attack: CoT mirage hacking.}
To validate our findings and exploit GPT-OSS-20B's procedure mirage, we design
a new attack. At a high level, the attack first constructs a benign request
similar to the harmful one (e.g., writing A Tale of Two Cities, whose copyright
is expired, instead of Harry Potter). The model's reasoning for the benign
request is then extracted, truncated, and modified through keyword replacement
with the original harmful context. Finally, the transformed chain of thought is
fed back into the target model to produce the harmful output. Detailed attack
process is described in Appendix~\cite{appx}.

\begin{figure}[t]
  \centering
  \includegraphics[width=0.6\linewidth]{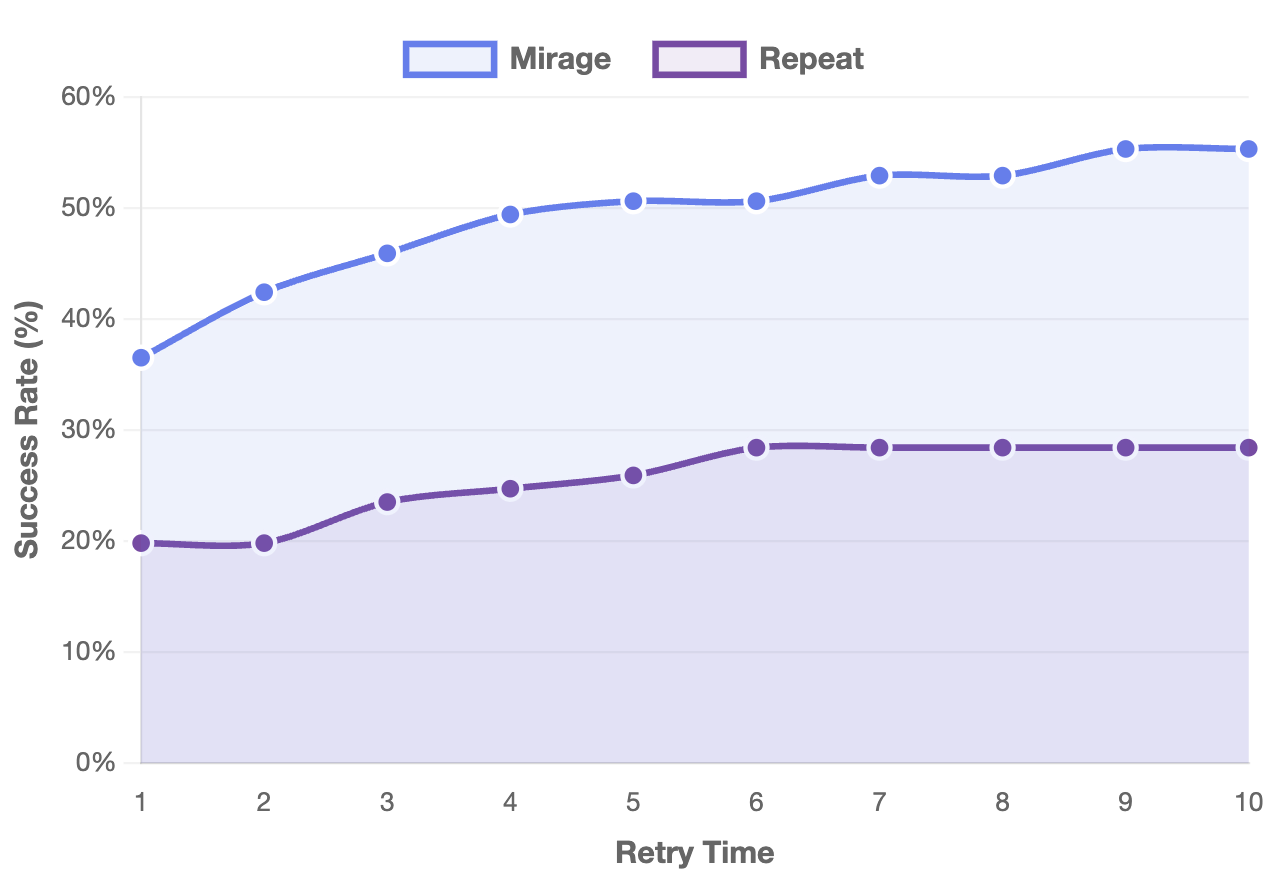}
  \caption{Comparison of the CoT mirage attack (our attack) with a
  content‑based CoT hijacking baseline~\cite{roe2025intriguing}.
  The mirage attack achieves higher jailbreak rates across prompts.}
  \label{fig:cot-mirage}
\end{figure}

Unlike content-based CoT hijacking~\cite{he2025removeharmony,bubbles2025cotoverride,reddit2025jailbreak},
our attack exploits the structure of CoT.
To validate its effectiveness, we compare against the CoT hijacking
baseline proposed by UChicago~\cite{roe2025intriguing}, which relies on repeated content-based
hijacking. Figure~\ref{fig:cot-mirage} shows the results---the gap between the two lines
reflects the effect of procedure mirage.

\subsection{Chain‑Oriented Prompting (COP)}
\label{ss:3.5}

We observe that in multi-round agentic scenarios, GPT-OSS-20B handles requests
in a locally valid manner without performing much of global checking, favoring
local coherence over long-term global safety. We exploit this behavior by
designing \emph{Chain-Oriented Prompting (COP)}. The core idea is inspired by
Return-Oriented Programming (ROP)~\cite{shacham2007geometry},
where attackers chain short instruction
sequences into an exploit. Similarly, COP decomposes a malicious goal into a
sequence of benign-looking prompts or tool calls. Each step appears safe in
isolation, but when executed in sequence, the agent assembles and performs a
harmful action. For example, while GPT-OSS-20B rejects a direct request to run
\texttt{rm -rf *}, an adversary can distribute the task across multiple prompts, leading
the model to unwittingly generate and execute the command.

\heading{COP in action.}
We evaluate COP in two attack scenarios: deleting files
(\texttt{rm -rf *}) and leaking private keys to the public domain. From our experiments,
GPT-OSS-20B does include some global checks---if adversarial prompts are simply
broken into four or five pieces and issued contiguously, the model refuses to
execute the final step. However, when we interleave the attack steps with
several noise prompts, COP succeeds much more frequently. In our experiments
(details in Appendix~\cite{appx}), we run each attack scenario 10 times and observe success
rates of 80\% and 70\%, respectively.

\section{Conclusion}

Our study shows that GPT-OSS-20B, though robust against standard harmful
prompts, exhibits unique and exploitable behaviors when systematically probed
with Jailbreak Oracle. We identify five key findings---quant fever, reasoning
blackholes, Schrodinger's compliance, reasoning procedure mirage, and
chain-oriented prompting (COP)---each enabling new attacks that bypass existing
safeguards by exploiting weaknesses in reasoning and alignment. We build attack
prototypes that raise jailbreak rates from 3.3\% to 44.4\% without CoT
manipulation (\S\ref{ss:3.3}), and from 28.4\% to 55.3\% with CoT manipulation (\S\ref{ss:3.4}). In
agentic settings, we further observe a ~70\% probability of risky behavior on
benign prompts with numerical targets (\S\ref{ss:3.1}), and a 70--80\% success rate for COP
attacks using prompt injection (\S\ref{ss:3.5}).
These findings highlight the urgent need
for defenses that address vulnerabilities arising not from isolated prompts,
but from compositional reasoning, procedural scaffolding, and adversarially
structured interactions.

\bibliography{ref}


\end{document}